\pdfoutput=1
\documentclass[11pt]{article}
\usepackage[normalem]{ulem}

\usepackage{NLPAICS2024}
\usepackage{times}
\usepackage{latexsym}
\usepackage{graphicx}
\usepackage{booktabs}
\usepackage{multirow}
\usepackage{arabtex}
\usepackage{utf8}
\usepackage{makecell}
\usepackage{amsmath}
\usepackage[T1]{fontenc}
\usepackage[utf8]{inputenc}

\usepackage{microtype}

\usepackage{tabularray}

\usepackage{amssymb}
\usepackage{pifont}

\title{DarijaBanking: A New Resource for Overcoming Language Barriers in Banking Intent Detection for Moroccan Arabic Speakers}

\author{
  Abderrahman Skiredj \\
  OCP Solutions and \\
  UM6P College of Computing, \\
  Casablanca, Morocco \\
  \texttt{abderrahman.skiredj@}\\ \texttt{ocpsolutions.ma} \\
  \And
  Ferdaous Azhari \\
  National Institute of \\
  Posts and Telecoms, \\
  Rabat, Morocco \\
  \texttt{azhari.ferdaous@}\\\texttt{doctorant.inpt.ac.ma} \\
  \AND
  Ismail Berrada \\
  UM6P College of Computing, \\
  Benguerir, Morocco \\
  \texttt{ismail.berrada@um6p.ma} \\
  \And
  Saad Ezzini \\
  School of Computing \\ and Communications \\
  Lancaster University, \\
  Lancaster, United Kingdom \\
  \texttt{s.ezzini@lancaster.ac.uk} \\
}

\begin{document}
\maketitle

\begin{abstract}
    Navigating the complexities of language diversity is a central challenge in developing robust natural language processing systems, especially in specialized domains like banking. The Moroccan Dialect (Darija) serves as the common language that blends cultural complexities, historical impacts, and regional differences. The complexities of Darija present a special set of challenges for language models, as it differs from Modern Standard Arabic with strong influence from French, Spanish, and Tamazight, it requires a specific approach for effective communication. To tackle these challenges, this paper introduces \textbf{DarijaBanking}, a novel Darija dataset aimed at enhancing intent classification in the banking domain, addressing the critical need for automatic banking systems (e.g., chatbots) that communicate in the native language of Moroccan clients. DarijaBanking comprises over 1,800 parallel high-quality queries in Darija, Modern Standard Arabic (MSA), English, and French, organized into 24 intent classes. We experimented with various intent classification methods, including full fine-tuning of monolingual and multilingual models, zero-shot learning, retrieval-based approaches, and Large Language Model prompting. One of the main contributions of this work is BERTouch, our BERT-based language model for intent classification in Darija. BERTouch achieved F1-scores of 0.98 for Darija and 0.96 for MSA on DarijaBanking, outperforming the state-of-the-art alternatives including GPT-4 showcasing its effectiveness in the targeted application.

\end{abstract}

\section{Introduction}

The field of Natural Language Processing (NLP) has gained significant traction, particularly with the emergence and advancement of Large Language Models (LLMs). This surge in interest and implementation is especially notable in industries where customer engagement and relationships are critical. In the realm of retail banking, Financial NLP and LLMs are reshaping the dynamics of client interactions, fostering enhanced personalization, responsiveness, and ultimately, customer loyalty and satisfaction \citep{ai_banking}. 

In the ecosystem of generative AI-enhanced customer service, powerful LLMs such as GPT-4 \citep{gpt4} act as the brain of the system, orchestrating various components to deliver nuanced and contextually relevant interactions \citep{agents}. Among these components, Retrieval Augmented Generation (RAG) \citep{rag} stands out by augmenting the agents' responses with information retrieved from a dense knowledge base, thereby enriching the quality and relevance of the interactions. However, the linchpin in this sophisticated setup is the \textbf{Intent Classification Module}, a critical aspect of Financial NLP's Natural Language Understanding (NLU).

Intent classification, also known as Intent Detection, focuses on deciphering the semantic essence of user inputs to elicit the most appropriate response. It involves creating a linkage between a user's request and the corresponding actions initiated by the chatbot, as outlined by \citet{adamopoulou}. Typically framed as a classification challenge, this process entails associating each user utterance with one or, in certain cases, multiple intended actions. The task of intent classification presents notable difficulties. Conversations with chatbots often involve terse utterances that offer minimal contextual clues for accurate intent prediction. Furthermore, the extensive variety of potential intents necessitates a vast dataset for annotation, complicating the detection process due to the expansive label space that must be managed. 

In this paper, we introduce DarijaBanking, a comprehensive Darija intent dataset and embark on a systematic comparison of diverse intent classification methodologies, encompassing full fine-tuning of both monolingual and multilingual models, zero-shot learning, retrieval-based strategies, and Large Language Model (LLM) prompting. DarijaBanking is meticulously curated from three foundational English banking datasets, namely:
\begin{itemize}
    \item Banking77 dataset \citep{banking77}, which provides a corpus of 13,083 banking-related queries each mapped to one of 77 distinct intents,
    \item Banking-Faq-Bot dataset \citep{banking_faq_bot}, which includes 1,764 questions distributed across 7 intent categories, and  
    \item Smart-Banking-Chatbot dataset \citep{smart_banking_chatbot}, encompassing 30,100 questions sorted into 41 intents, cumulatively aggregating to a rich repository of 42,000 questions.
\end{itemize}

The initial phase involved a rigorous cleaning process aimed at intent unification across these diverse sources, addressing various challenges such as generic or colloquially dense sentences, redundancy, and contextual incompatibilities with the Moroccan banking sector. This meticulous refinement resulted in a distilled collection of 1,800 English sentences, subsequently translated into French and Modern Standard Arabic (MSA) using the OPUS MT \citep{opus_mt} and Turjuman \citep{turjuman} models, respectively. Subsequently, a crucial translation step involved leveraging GPT-4 \citep{gpt4} to translate these English sentences into Darija.

To ensure the highest level of accuracy and contextual relevance, particularly for the Darija translations, we employed a team of five external human native speakers as annotators. The annotators conducted a thorough manual verification and correction process, focusing on the nuanced aspects of the Moroccan dialect that automated systems might overlook. Their invaluable contributions were instrumental in refining the dataset to accurately reflect the linguistic intricacies of Darija, thereby enhancing the dataset's utility for our intended applications. It is noteworthy that all utterances underwent manual review, with approximately 47\% of them were edited to further ensure their accuracy and idiomatic appropriateness.

The resultant dataset, featuring a total of 7,200 queries across English, French, MSA, and Darija comprising of 1,800 queries for each language. The MSA and Darija subsets serve as the foundation for assessing various intent classification methods, including full fine-tuning of monolingual and multilingual models, zero-shot learning, retrieval-based approaches, and Large Language Models prompting. 

The main contributions of our paper can be succinctly summarized as follows:

\begin{itemize}
    \item The \textbf{DarijaBanking dataset} is introduced, serving as a novel resource for banking intent detection in Darija, featuring over 7,200 multilingual queries across English, French, MSA, and Darija across 24 intent categories. This dataset results from arabizing, localizing, deduplicating, cleansing, and translating existing English banking data.
    
    \item A \textbf{comparative analysis} of intent classification methodologies is presented, evaluating monolingual and multilingual model fine-tuning, zero-shot learning, retrieval strategies, and LLM prompting, to assess their effectiveness across the dataset.

    \item \textbf{BERTouch} our Darija specific BERT-based language model, that is tailored for the intent classification task. BERTouch proved to be competitive with state-of-the-art solution including large language models such as GPT-4. As a contribution to open science, we made BERTouch publicly available in the HuggingFace platform.

    \item \textbf{Insights for enhancing Moroccan Darija banking intent detection systems} are detailed, highlighting the balance between using generalist LLMs, cross-lingual transfer learning, and the importance of domain-specific data annotation. Our findings indicate the advantage of specialized classifiers with sufficient data labeling and the efficacy of retrieval-based approaches as a budget-friendly option, guiding the development of precise and economical systems.
\end{itemize}

The rest of the paper is structured to provide a holistic view of this research endeavor. Section 2 delves into the related work, establishing a contextual backdrop. Section 3 unveils the DarijaBanking corpus, detailing the processes of data arabization and localization. Section 4 explores the array of intent classification approaches, including model architectures and training paradigms. Section 5 showcases the empirical results derived from these methodologies. Section 6 is dedicated to a thorough discussion of the results obtained from our experiments, drawing critical insights and identifying the implications of our findings for the field, while also discussing the limitations of our work. Finally, we conclude our paper by summarizing the key takeaways, reflecting on the contribution of the DarijaBanking corpus to the advancement of intent detection in the Moroccan Dialect.

\section{Related Work}

In the realm of NLP, the Arabic language presents unique challenges, primarily due to its rich dialectical diversity and the scarcity of domain-specific labeled datasets. This scarcity significantly impedes the advancement of Arabic NLP applications, including intent detection and conversational systems. Recent research endeavors, as documented by \citet{Darwish_et_al_2021} and \citet{Naser_Karajah_et_al_2021}, highlight the acute shortage of labeled datasets for Arabic, especially for dialectal and specialized tasks. Similarly, \citep{Fuad_and_Al_Yahya_2022} and \citep{Ahmed_et_al_2022} underscore the resultant stagnation in the development of Arabic conversational machine learning systems, attributing this lag to the inadequate dataset resources. Furthermore, there has been a recent surge in interest in Arabic NLP, as evidenced by the shared task \citep{arafinnlp2024}, which aims to foster further research and development in this area.

Despite these challenges, some strides have been made towards understanding and processing the Arabic language more effectively. A pioneering step in the domain of Arabic intent detection was taken by \citep{Mezzi_et_al_2022}, who introduced an intent detection framework tailored for the mental health domain in Tunisian Arabic. Their innovative approach involved simulating psychiatric interviews using a 3D avatar as the interviewer, with the conversations transcribed from audio to text for processing. The application of a BERT encoder and binary classifiers for each mental health aspect under investigation—depression, suicide, panic disorder, social phobia, and adjustment disorder—yielded impressive results, achieving an F1 score of 0.94.

Exploring other facets of Arabic NLP, \citep{Hijjawi_et_al_2013} ventured into the classification of questions and statements within chatbot interactions, leveraging decision trees for this task. Their methodology was later integrated into ArabChat \citep{Hijjawi_et_al_2014}, enhancing the system's ability to preprocess and understand user inputs. Moreover, \citep{Joukhadar_et_al_2019} contributed to the field by creating a Levantine Arabic corpus, annotated with various communicative acts. Their experiments with different classifiers revealed that the Support Vector Machine (SVM), particularly when utilizing 2-gram features, was most effective, achieving an accuracy rate of 0.86.

The quest for understanding Arabic speech acts and sentiments led \citet{Elmadany_et_al_2018} to develop the ArSAS dataset, encompassing around 21K tweets labeled for speech-act recognition and sentiment analysis. The dataset, marked by its categorization into expressions, assertions, questions, and sentiments (negative, positive, neutral, mixed), provided a fertile ground for subsequent studies. Utilizing ArSAS, \citet{Algotiml_et_al_2019} employed Bidirectional Long-Short Term Memory (BiLSTM) and SVM to model these nuanced linguistic features, achieving an accuracy of 0.875 and a macro F1 score of 0.615. Lastly, \citet{Zhou_et_al_2022} demonstrated the potential of contrastive-based learning to enhance model performance on out-of-domain data, testing their methodology across several datasets, including the banking domain \citep{banking77}, and showed that it's possible to improve adaptability without sacrificing the accuracy for in-domain data.

Research on intent detection has progressed beyond Arabic, with notable studies in languages like Urdu and Indonesian. \citet{Shams_et_al_2019} translated key datasets to Urdu and employed CNNs, LSTMs, and BiLSTMs, finding CNNs most effective for ATIS with a 0.924 accuracy and BiLSTMs best for AOL at 0.831 accuracy. This work was further refined to achieve a 0.9112 accuracy \citep{Shams_and_Aslam_2022}. Similarly, in Indonesian, \citet{Bilah_et_al_2022} utilized ATIS to inform their CNN model, achieving a 0.9584 accuracy.

Expanding the scope, \citet{Basu_et_al_2022} explored a meta-learning approach with contrastive learning on Snips and ATIS datasets for diverse domains, emphasizing the complexity of intent detection across different contexts.

Addressing the gap in Arabic intent detection, \citet{arbanking} introduced ArBanking77, an arabized dataset from Banking77 \citep{banking77}, enhanced with MSA and Palestinian dialect queries, totaling 31,404 queries across 77 intents. A BERT-based model finetuned on this dataset achieved F1-scores of 0.9209 for MSA and 0.8995 for the Palestinian dialect.

In the exploration of NLP for Moroccan Darija, significant strides have been made across two pivotal areas: the development of novel datasets and the advancement in classification tasks. The creation of specialized datasets, as seen in the work of \citet{essefar}, introduces the Offensive Moroccan Comments Dataset (OMCD), marking a significant step towards understanding and moderating offensive language in Moroccan Arabic. This initiative fills a gap in resources tailored to the nuances of Moroccan dialect. Similarly, \citet{boujou}'s contribution of a multi-topic and multi-dialect dataset from Twitter, which includes Moroccan Darija, provides a versatile tool for sentiment analysis, topic classification, and dialect identification, underlining the importance of accessible data for diverse NLP applications.
On the classification front, \citet{mekki}'s research leverages a deep Multi-Task Learning model with a BERT encoder for nuanced language identification tasks. This model's ability to differentiate between various Arabic dialects and MSA showcases the evolving precision in language processing techniques. Furthermore, \citet{adasl} introduces AdaSL, an unsupervised domain adaptation framework aimed at enhancing sequence labeling tasks such as Named Entity Recognition and Part-of-Speech tagging. Their approach, which capitalizes on unlabeled data and pre-trained models, addresses the challenges posed by the scarcity of labeled datasets in Dialectal Arabic, thereby advancing the field of token and sequence classification within Moroccan Darija and other dialects.

Building on these contributions, our study specifically targets the unexplored area of intent detection for Moroccan Darija within the banking domain. We introduce DarijaBanking, a pioneering dataset crafted to enhance intent detection capabilities. This dataset, developed through the arabization, localization, deduplication, thorough cleaning and translation of existing English banking datasets, comprises over 3,600 queries across 24 intent classes, in both Modern Standard Arabic and Moroccan Darija. In exploring various methods for intent detection, we highlight the success of a BERT-based model specifically finetuned for this dataset, which achieved impressive F1-scores of 0.98 for Darija and 0.96 for MSA.

\section{The DarijaBanking Corpus}

The DarijaBanking Corpus\footnote{https://github.com/abderrahmanskiredj/DarijaBanking} represents a novel endeavour to tailor and enrich banking-related linguistic resources specifically for the Moroccan context, leveraging the foundational structures of three significant English datasets, namely: (1) Banking77 \citep{banking77},  which encompasses an extensive collection of 13,083 banking-related queries, each meticulously categorized into one of 77 distinct intents; (2) the banking-faq-bot dataset \citep{banking_faq_bot}, comprising 1,764 questions distributed across seven intent categories; and (3) the smart-banking-chatbot dataset \citep{smart_banking_chatbot}, that includes a broad spectrum of 30,100 questions sorted into 41 intents. Collectively, these resources amalgamate into a comprehensive repository of 42,000 questions.
Subsequently, we will detail the various stages involved in our corpus's data collection and validation process.

\subsection{Data Collection}
The Data Collection comprises 4 phases:

\noindent
\paragraph{Phase I: Cleaning.} The first step in developing the DarijaBanking corpus was a rigorous cleaning process tailored to align the dataset with the nuances of Morocco's banking sector. This essential phase focused on eliminating queries and intents associated with banking practices and products common in countries like the US or UK but irrelevant or nonexistent in Morocco. We aimed to exclude references to unfamiliar banking services within the Moroccan banking environment, such as specific types of loans, investment opportunities, or account functionalities unavailable in local banks. For instance, intents related to "Apple Pay or Google Pay," "Automatic Top-Up," "Disposable Card Limits," "Exchange Via App," "Get Disposable Virtual Card," "Topping Up by Card," and "Virtual Card Not Working" were removed due to their limited relevance to Moroccan banking users. This is because the penetration of digital wallet services such as Apple Pay and Google Pay is not as extensive in Morocco, making these services less applicable. Additionally, the concept of automatic top-ups, disposable virtual cards, and app-based foreign exchange are indicative of fintech advancements not fully adopted or supported by Moroccan banking institutions, which tend to offer more traditional services. The use of virtual cards and card-based account top-ups, while increasing worldwide, might not yet align with common banking practices or the digital infrastructure in Morocco.

The cleaning also involved a critical step of utterance-level filtering to bolster the corpus's relevance to the Moroccan banking context, by eliminating references to:
\begin{itemize}
\item Transactions linked to UK bank accounts, e.g., "I made a transfer from my UK bank account."
\item Queries on services for non-residents, e.g., "Can I get a card if I'm not in the UK?"
\item The use of foreign currencies and credit cards, e.g., "I need Australian dollars," "Can I use a US credit card to top up?"
\item International transfers and related issues, e.g., "How long does a transfer from the UK take?" "My transfer from the UK isn't showing."
\end{itemize}
Removing utterances involving international banking contexts, uncommon currency conversions in Morocco, or services related to foreign accounts made the corpus more reflective of Moroccan banking scenarios.

A subsequent challenge was addressing ambiguous intent clusters that hindered clear intent detection. Clusters such as "card not working" and "compromised card" exemplified the issue, with the former combining intents like "activate\_my\_card" and "declined\_card\_payment," and the latter grouping "compromised\_card" with "lost\_or\_stolen\_card." Similarly, "transfer problems" and "identity verification" clusters highlighted the difficulty in distinguishing between closely related intents, such as "failed\_transfer" and "transfer\_not\_received\_by\_recipient," or "unable\_to\_verify\_identity" and "verify\_my\_identity," respectively. These clusters, among others, demonstrated the substantial overlap and frequent misclassification of intents, complicating the dataset's utility for accurate intent recognition.
To mitigate this, a strategic decision was made to refine the dataset by merging similar intents and eliminating those prone to classification errors. This consolidation aimed to simplify the intent detection process, enhancing the model's accuracy by focusing on broader, more distinguishable intent categories.

Furthermore, deduplication was a key step in refining the DarijaBanking corpus, targeting nearly identical queries with slight wording differences but the same intent, such as inquiries about SWIFT transfers or using Apple Pay. This step improved the dataset's precision, aiding the intent detection model in more accurately classifying customer intents in Morocco's banking sector.

Through detailed refinement, including deduplication and correction of incorrect utterance-intent associations, we developed a polished collection of 1,660 English sentences, significantly enhancing the dataset's utility for accurate intent recognition within the Moroccan banking context.

\paragraph{Phase II: The two additional intents IDOOS and OODOOS.}
In the subsequent phase, we expanded the DarijaBanking corpus by incorporating two additional intent categories: In-Domain Out-Of-Scope (IDOOS) and Out-Of-Domain Out-Of-Scope (OODOOS). IDOOS encompasses requests that, while not previously listed, remain pertinent to the banking sector, such as inquiries about Western Union facilities. By adding IDOOS, we ensure the chatbot can recognize and manage banking-related queries that fall outside the predefined set of intents. This expansion allows the bot to cater to a broader array of banking inquiries, enhancing its utility and user satisfaction by reducing instances where legitimate banking questions go unrecognized.
Conversely, OODOOS covers requests that are entirely unrelated to banking, exemplified by questions regarding the distance between the Earth and the Moon. The inclusion of OODOOS helps the chatbot identify and appropriately handle queries unrelated to banking. This distinction is crucial for maintaining the chatbot's focus on banking topics and directing users to suitable resources for their non-banking questions, thereby improving the overall user experience by preventing irrelevant responses.
For each of these new intent categories, we integrated 70 English utterances generated by ChatGPT, enriching the corpus to encompass a total of 1,800 English sentences across 24 distinct intents.
Examples of these intents include:\\
\textbf{IDOOS Examples:}
\setcode{utf8}
\begin{itemize}
    \item Are there any blackout dates or restrictions on redeeming my rewards?
    \item Can you explain the different types of retirement accounts?
    \item (How do I learn to invest in the stock market?) $$\RL{كيفاش نتعلم نستثمر فالبورصة؟} $$
    \item (How do I use health savings account funds to invest?)
    \begin{center}
    \RL{كيفاش نستعمل أموال حساب}\\
    \RL{ توفير الصحة باش نستثمر؟}\\
    \end{center}
    \item (How do I earn reward points with my card?) 
    \begin{center}
    \RL{كيفاش ندير باش نكسب نقاط}\\
    \RL{ المكافآت ديال البطاقة ديالي؟}\\
    \end{center}
\end{itemize}

\textbf{OODOOS Examples:}
\begin{itemize}
    \item (What is the average lifespan of a cat?) $\RL{شنو هي المدة المتوسطة ديال عمر القطة؟} $
    \item  (What's the best way to learn a new language?) $\RL{شنو هيا أحسن طريقة باش تعلم لغة جديدة؟}$
    \item (Who won last year's World Cup?) $\RL{شكون ربح كأس العالم العام لي فات؟} $
\end{itemize}

\paragraph{Phase III: Automatic Translation}
In this phase, the meticulously cleaned English sentences were translated into French, Arabic, and Darija employing the OPUS MT \citep{opus_mt}, Turjuman \citep{turjuman}, and GPT-4 \citep{gpt4} models, respectively, for this purpose.

\paragraph{Phase IV: Manual Verification \& Correction}

Here the emphasis was placed not only on linguistic precision but also on ensuring that the translations were contextually aligned with Moroccan banking terminology. The endeavor went beyond mere translation to include the crucial aspect of localization. For instance, the term "transfer" in a banking context was appropriately translated not as $\RL{نقل}$, which conveys a general sense of moving or transferring, but as $\RL{تحويل}$, accurately capturing the notion of financial transfers. This understanding of banking terminology and its correct application is vital for the dataset's relevance and utility. Additional examples include "savings account," which required careful translation to $\RL{حساب التوفير}$ to avoid a generic rendition, and "loan approval," which was precisely translated to $\RL{الموافقة على القرض}$, eschewing a less appropriate literal translation. These instances highlight the critical need for a nuanced approach to translation and localization, ensuring the dataset's resonance with the intricacies of Moroccan banking language and operations.

The process of manual correction and verification of the translations from English to Moroccan Darija was meticulously conducted by a team of \textbf{five human labelers} who are native speakers of Moroccan Darija. \textbf{Each translated utterance was manually checked, resulting in approximately 47\% of them being edited to enhance accuracy and idiomatic expression}. These labelers reviewed the initial translations provided by GPT-4 and made necessary adjustments to ensure that the translations were accurate and idiomatic to Moroccan Darija speakers.
The types of corrections that were performed included standardizing terms, correcting verb conjugations, and ensuring the use of dialectally appropriate vocabulary. For example, the initial translation for "Will deactivating my card affect my automatic bill payments?" was
\begin{eqnarray*}
    &\RL{غادي تعطيل بطاقتي يأثر}\\
    &\RL{ في الدفعات التلقائية ديالي؟}
\end{eqnarray*}
%\begin{center}
 %   \RL{غادي تعطيل بطاقتي يأثر}\\
  %  \RL{ في الدفعات التلقائية ديالي؟}
%end{center}
which was corrected to 
\begin{center}
    \RL{واش تعطيل بطاقتي غادي يأثر}\\
    \RL{ في الدفعات التلقائية ديالي؟}
\end{center}
by replacing the initial word $\RL{غادي}$ with $\RL{واش}$ to better fit a question structure in Moroccan Darija. Some more examples are illustrated in table \ref{manual_correction}
In broader terms, the corrections made sure that the translations were not only linguistically accurate but also culturally resonant, taking into account the nuances of Moroccan Darija and its syntactic and lexical particularities. This quality control process is crucial in ensuring that translations are not only understood but also feel natural to Moroccan Darija speakers.

\setcode{utf8}
% \usepackage{graphicx}
\begin{table*}
\centering
\resizebox{\textwidth}{!}{%
\begin{tabular}{|l|l|}
\hline
\multicolumn{1}{|c|}{\textbf{GPT4   Translation To Darija}}                    & \multicolumn{1}{c|}{\textbf{Correction Translation To Darija}}            \\ \hline
$\RL{كنقدرو ننصحوك   بمستشار مالي جيد؟}$                                & $\RL{تقدر تنصحني بمستشار مالي مزيان؟} $                             \\ \hline
$\RL{واش كيمكنني نخدم   قرض باش نسافر بيه؟} $                            & $\RL{واش يمكن ليا ناخد قرض باش نسافر بيه؟}$                         \\ \hline
$\RL{كنت مفيد جدا :)}$                                                   & $\RL{كنتي مفيد بزاف :)}$                                            \\ \hline
$\RL{هل ممكن تسول وكيلة   إذا نقدر نجدد معلومات فبروفيلي؟}$              & $\RL{ممكن تسول شي وكيل إيلاا نقدر نجدد معلومات   فبروفيلي؟}$        \\ \hline
$\RL{مساعدة! تنيت   كارتي!}$                                             & $\RL{عاونوني ! ضاعت ليا البطاقة ديالي }$                            \\ \hline
$\RL{فتوري الأخيرة،   أخذت معامل صرف غلط.}$                              & $\RL{فالرحلة الأخيرة ديالي، خذيت معامل صرف غلط.}$                   \\ \hline
$\RL{بغيت نهدر مع واحد   الشخص }$                                        & $\RL{بغيت نهدر مع شي واحد الشخص }$                                  \\ \hline
$\RL{احتاج تأيد باش   نقفل حسابي.}$                                      & $\RL{بغيت مساعدة باش نقفل حسابي.}$                                  \\ \hline
$\RL{غلطو فسعر الصرف   بوقت اشتريت حاجة فبلاد خارجية.}$                  & $\RL{غلطو فسعر الصرف فاش شريت حاجة من برا}$                         \\ \hline
$\RL{بغيت نعرف أيش هو   الرقم ديال خدمة الزبون }$                        & $\RL{بغيت نعرف أشنو هو الرقم ديال خدمة الزبون }$                    \\ \hline
$\RL{قادر تهدرني على   العملية ديال إلغاء طلبية لي تدفعت؟ }$             & $\RL{تقدر توجهني إلى العملية ديال إلغاء طلبية لي تدفعت؟   }$        \\ \hline
$\RL{غيرت راي، شنو   الدبا لي عندي أندير باش نسكر حساب المستعمل ديالي؟}$ & $\RL{غيرت راي، شنو خاصني دبا ندير باش نحيد حساب   المستعمل ديالي؟}$ \\ \hline
\begin{tabular}[c]{@{}l@{}}$\RL{شنو يجب ندير إذا   كنت كتشكي أن واحد وجد تيليفوني}$\\  $\RL{لي ضاع أو سرق و كيعمل بيه معاملات غير مصرح   بها؟}$\end{tabular} &
  \begin{tabular}[c]{@{}l@{}}$\RL{شنو خاص نديرو اذا كنت كانشك فشي واحد لقا}$\\  $\RL{  تيليفوني  لي ضاع أو سرق و كيعمل بيه   معاملات غير مصرح بها؟}$\end{tabular} \\ \hline
$\RL{أي رسوم يتطبقوا   لما كنستعمل كارت؟}$                               & $\RL{أشمن رسوم كيتطبقوا فاش كنستعمل لاكارت} $                       \\ \hline
$\RL{شنو هو التفرق بين   فرشاة الأسنان العادية والكهربائية؟} $           & $\RL{شنو هو الفرق بين فرشاة الأسنان العادية   والكهربائية؟}$        \\ \hline
$\RL{واش نقدر نحصَّل   ردّة؟}$                                           & $\RL{واش نقدر ناخد تعويض؟}$                                         \\ \hline
$\RL{كيفاش نعطل بطاقتي   إذا كنتشفق أنها تعرضت لخطر؟ }$                  & $\RL{كيفاش نعطل بطاقتي إذا شكيت أنها تعرضت لخطر؟ }$                 \\ \hline
$\RL{ما عنديش هاتفي   بعد، شنو يمكن ندير؟}$                              & $\RL{ما بقاش عندي تيليفون، شنو يمكن ندير؟}$                         \\ \hline
$\RL{فيه سن معينة لازم   نكون عندها باش نستعملوا خدماتكم؟ }$             & $\RL{شنا هو العمر المطلوب باش نفتح  حساب؟}$                         \\ \hline
\end{tabular}%
}
\caption{Some examples of manually corrected translations from English to Darija}
\label{manual_correction}
\end{table*}

\subsection{Comprehensive Intent Catalogue}

Table \ref{catalogue} delineates a comprehensive list of intents featured in the corpus, along with their corresponding definitions. The detailed overview serves to illustrate the dataset's scope concerning customer interactions and inquiries specific to the banking industry. In total, the corpus comprises 22 intent, plus the additional IDOOS and OODOOS as described above. The number of utterances per intent ranges from 58 to 99, with an average of 74, indicating a well-balanced distribution across the dataset.

% \usepackage{graphicx}
\begin{table*}[!htb]
\centering
\resizebox{\textwidth}{!}{%
\begin{tabular}{|l|l|c|}
\hline
\multicolumn{1}{|c|}{\textbf{Intent}} &
  \multicolumn{1}{c|}{\textbf{Description}} &
  \textbf{\begin{tabular}[c]{@{}c@{}}Nb of utterances\\ by intent\\ for each language\end{tabular}} \\ \hline
activate\_my\_card          & Initiate the use of a new banking card.                    & 64 \\ \hline
age\_limit                  & Inquire about the minimum age requirement for a service.   & 58 \\ \hline
cancel\_order               & Request to cancel a previously placed order.               & 74 \\ \hline
cancel\_transfer            & Request to cancel a previously initiated money transfer.   & 66 \\ \hline
card\_swallowed             & Report an ATM machine retaining a banking card.            & 82 \\ \hline
change\_order               & Modify details of a previously placed order.               & 89 \\ \hline
compromised\_card           & Report a banking card suspected of being at risk of fraud. & 75 \\ \hline
contact\_customer\_service  & Request for assistance from the bank’s customer service.   & 96 \\ \hline
contact\_human\_agent       & Seek to speak with a live customer support agent.          & 94 \\ \hline
create\_account             & Initiate the process of opening a new bank account.        & 96 \\ \hline
deactivate\_my\_card        & Disable a currently active banking card.                   & 64 \\ \hline
delete\_account             & Request the closure of a bank account.                     & 95 \\ \hline
edit\_account               & Make changes to the account information.                   & 71 \\ \hline
exchange                    & Inquire about currency exchange services.                  & 76 \\ \hline
fee                         & Question about the charges associated with a service.      & 64 \\ \hline
general\_negative\_feedback & Provide negative feedback on the overall service.          & 68 \\ \hline
general\_positive\_feedback & Provide positive feedback on the overall service.          & 64 \\ \hline
get\_invoice                & Request a bill or invoice for a transaction.               & 99 \\ \hline
get\_refund                 & Request a return of funds for a transaction.               & 91 \\ \hline
insurance                   & Inquire about insurance products offered by the bank.      & 63 \\ \hline
loan                        & Request information on loan products.                      & 62 \\ \hline
lost\_or\_stolen\_phone     & Report a lost or stolen phone linked to mobile banking.    & 61 \\ \hline
idoos &
  \begin{tabular}[c]{@{}l@{}}An intent not in the list of intents but within the banking\\  domain, like asking for Western Union facilities.\end{tabular} &
  71 \\ \hline
oodoos &
  \begin{tabular}[c]{@{}l@{}}An intent not in the list of intents and not related to banking,\\  like asking the distance between the Earth and the Moon.\end{tabular} &
  64 \\ \hline
\end{tabular}%
}
\caption{Comprehensive Intent Catalogue}
\label{catalogue}
\end{table*}

\subsection{Data Segmentation and Descriptive Analysis}

The final dataset emerged as an extensive collection, comprising 7,200 queries distributed among English, French, Arabic, and Darija languages, with a concentrated segment of 3,600 queries specifically in Arabic and Darija, spanning across 24 distinct intents. The division of the dataset for training and testing purposes followed an 80:20 ratio, respectively, opting not to designate a separate validation subset due to the manageable size of the dataset. This stratified split was meticulously designed to maintain the original proportion of utterances per intent, ensuring that these ratios were consistently mirrored in both the training and testing subsets. Table \ref{stats} provides detailed statistics on the DarijaBanking corpus, offering insights into the dataset's composition and the distribution of utterances among the various intents.

% \usepackage{graphicx}
\begin{table}[h]
\centering
\resizebox{\columnwidth}{!}{%
\begin{tabular}{|c|c|c|c|c|c|}
\hline
\textbf{} & \textbf{English} & \textbf{French} & \textbf{MSA} & \textbf{Darija} & \textbf{Overall} \\ \hline
\textbf{Query count}       & 1800 & 1800 & 1800 & 1800 & 7200 \\ \hline
\textbf{Avg word count}    & 10,5 & 10,2 & 9,5  & 7,9  & 8,7  \\ \hline
\textbf{Min word count}    & 1    & 1    & 1    & 1    & 1    \\ \hline
\textbf{Max word count}    & 56   & 54   & 49   & 36   & 49   \\ \hline
\textbf{Std of word count} & 5,08 & 5,13 & 4,44 & 3,77 & 4,2  \\ \hline
\end{tabular}%
}
\caption{Statistics of DarijaBanking dataset}
\label{stats}
\end{table}

\section{Intent Detection Approaches}
In this section, we embark on a systematic comparison of diverse methodologies for intent detection, which can broadly be categorized into three distinct approaches: BERT-like models finetuning, retrieval-based strategies, and LLM prompting. The finetuning of BERT-like models \cite{devlin-etal-2019-bert} includes both monolingual and multilingual variations, where we specifically focus on adapting monolingual Arabic models for Arabic and Moroccan Darija, and employ cross-lingual transfer techniques with multilingual models like XLM-RoBERTa \citep{conneau-etal-2020-unsupervised} across English, French, Arabic, and Darija datasets. Additionally, we explore zero-shot learning scenarios to evaluate model performance on languages or dialects not explicitly included in the training phase. Meanwhile, the retrieval-based strategy leverages text embedding models to index and match queries to the nearest utterance, thus inferring intent based on semantic similarity. Lastly, LLM prompting involves the utilization of advanced models such as ChatGPT-3.5 \cite{chatgpt} and GPT-4 \citep{gpt4}, which are prompted to classify intents by providing a comprehensive list of intents and their descriptions. Each of these approaches offers a unique perspective on intent detection, highlighting the versatility and adaptability of current technologies to understand and classify user intents across a range of languages and contexts.

\subsection{BERT-like Models Finetuning}\label{intent_detection_by_finetuning}

The BERT model, developed by Google in 2018, has revolutionized the field of NLP by introducing a transformer-based architecture that excels in a wide range of common language tasks, including sentiment analysis, named entity recognition, and question answering \citep{devlin-etal-2019-bert}. The core of BERT's architecture is the transformer, utilizing an attention mechanism that efficiently learns contextual relationships between words in a given sequence. This architecture comprises two primary components: an encoder that processes the input text, and a decoder that generates predictions for the task at hand, such as masked token prediction or next-sentence prediction. This innovative approach has enabled BERT to achieve state-of-the-art performance across various NLP benchmarks.
In the context of this paper, we finetune a pre-trained BERT model for intent detection. To adapt BERT for the intent classification task, a single linear layer is appended to the pre-existing transformer layers of the BERT model. This modification allows for the direct application of BERT's contextual embeddings to the task of intent detection, leveraging its understanding of language nuances to achieve high accuracy.

Given the linguistic diversity and complexity of the Arabic language and its dialects, the performance of multilingual pre-trained transformers, including BERT, often varies. Recognizing this challenge, researchers have developed several BERT-like models tailored to Arabic and its dialects. These models have been pre-trained on vast corpora of Arabic text, encompassing both MSA and various regional dialects, to capture the rich linguistic features unique to Arabic.
Among these models, AraBERT \citep{arabert} stands out, having been trained on substantial Arabic datasets, including the 1.5 billion-word Abu El-Khair corpus and the 3.5 million-article OSIAN corpus. Similarly, ARBERT and MARBERT \citep{arbert}, as well as MARBERTv2, have been trained on extensive collections of MSA and dialectical Arabic texts, with MARBERTv2 benefiting from even more data and longer training regimes. QARiB \citep{qarib} represents another significant contribution, developed by the Qatar Computing Research Institute (QCRI) and trained on a diverse mix of Arabic Gigaword, Abu El-Khair Corpus, and Open Subtitles. Lastly, CAMeLBERT-Mix \citep{camelbert} incorporates a broad spectrum of MSA and dialectical Arabic sources, including the Gigaword Fifth Edition and the OpenITI corpus, to create a model that is well-suited for a wide range of Arabic NLP tasks.
Building on the landscape of BERT-like language models, XLM-RoBERTa \citep{conneau-etal-2020-unsupervised} emerges as a pivotal development, extending the capabilities of language understanding beyond the realm of single-language models. XLM-RoBERTa is architected on the robust foundation of RoBERTa \citep{roberta}, leveraging a Transformer-based framework to foster deep contextual understanding across languages. Trained on an expansive corpus encompassing texts from 100 different languages, XLM-RoBERTa utilizes a large-scale approach without the need for parallel corpora, focusing on the Masked Language Model (MLM) objective to predict randomly masked tokens within a text for profound contextual comprehension. Additionally, it incorporates the Translation Language Model (TLM) objective in specific training setups, where it learns to predict masked tokens in bilingual text pairs, further enhancing its cross-lingual capabilities.
In this paper, we evaluate these Arabic pre-trained transformer models, alongside the multilingual XLM-Roberta \citep{conneau-etal-2020-unsupervised} on the DarijaBanking dataset.

\subsection{Retrieval-based Intent Detection}\label{intent_detection_by_retrieval}

Retrieval-based intent detection represents a pragmatic approach towards understanding and classifying user intents. By employing sophisticated text embedding models, each utterance within the training dataset is transformed into a dense vector representation. When a new query is received, it is also embedded into the same vector space. The intent of the query is inferred by identifying the nearest utterance in the training set, where "nearest" is quantified by the highest cosine similarity between the query's embedding and those of the dataset utterances. This method hinges on the premise that semantically similar utterances, when properly embedded, will occupy proximate regions in the embedding space, facilitating an efficient and effective retrieval-based classification.

\paragraph{Neural Architectures for Text Embedding Models.} Text embedding models, particularly those inspired by the BERT architecture, have been central to advancements in NLP. However, traditional BERT models do not directly compute sentence embeddings, which poses challenges for applications requiring fixed-length vector representations of text. A common workaround involves averaging the output vectors or utilizing the special CLS token's output, though these methods often result in suboptimal sentence embeddings. To address these limitations, Sentence-BERT \citep{sentence-bert} was introduced as an adaptation of the BERT framework, incorporating siamese and triplet network structures to generate semantically meaningful sentence embeddings. These embeddings can then be compared using cosine similarity, offering a significant boost in efficiency and effectiveness across various sentence-level tasks.

\paragraph{Multilingual Text Embedding Models Supporting Arabic.}
Several models have emerged to support multilingual and dialect-specific embeddings, crucial for applications involving Arabic and its dialects:

\begin{itemize}

\item Sentence Transformers and Knowledge Distillation: The approach taken by models such as the one described by \citep{distiluse} combines the strength of Sentence Transformers with the concept of knowledge distillation. Here, a monolingual "teacher" model guides a "student" model to produce comparable sentence embeddings across languages, facilitated by training on parallel corpora. Within this family of models, several have been specifically considered for their ability to handle Arabic. These include distiluse-base-multilingual-cased-v1 \citep{distiluse_hf} which operates within a 512-dimensional dense vector space; paraphrase-multilingual-mpnet-base-v2 \citep{paraphrase_hf}, offering embeddings in a 768-dimensional space; and paraphrase-multilingual-MiniLM-L12-v2 \citep{minilm_hf}, which provides embeddings in a 384-dimensional dense vector space.

\item LaBSE (Language-Agnostic BERT Sentence Embedding): Google's LaBSE model \citep{labse} represents a significant leap forward, blending techniques for monolingual and cross-lingual embedding generation. By integrating masked language modeling, translation language modeling, and other advanced methods, LaBSE achieves robust multilingual and dialectal coverage. LaBSE offers embeddings in a 768-dimensional space.

\item LASER (Language-Agnostic SEntence Representations): Developed by Meta, LASER \citep{LASER} employs a BiLSTM encoder with a shared BPE vocabulary across languages. This model excels in generating language-agnostic sentence embeddings, supported by extensive parallel corpora training. LASER offers embeddings in a 1024-dimensional space.

\item E5 Model: The E5 model \citep{e5} introduces a contrastive training approach with weak supervision, yielding embeddings that excel in retrieval, clustering, and classification tasks across languages, including zero-shot and finetuned scenarios. Within this family of models, the base model with its 12 layers was used. It provides embeddings in a 768-dimensional space.

\item OpenAI's text-embedding-3-large \citep{openai_embedding}: This model represents OpenAI's latest advancement in text embedding technology. It's designed for high performance and flexibility, offering embeddings with up to 3072 dimensions. This model is particularly notable for its enhanced multilingual performance, making it a valuable tool for tasks requiring a nuanced understanding of text across various languages.

\end{itemize}
Despite the advancements in text embedding models and their application to a multitude of languages, it is important to acknowledge a significant limitation when it comes to handling Arabic and its diverse dialects. The aforementioned models, while multilingual and capable of supporting Arabic to a degree, are not specifically designed with Arabic and its dialects as a primary focus. Their training datasets predominantly encompass MSA, with limited or, in some cases, no exposure to the various dialects spoken across the Arab world. This gap underscores a critical challenge in the field of NLP: the development of embedding models that can accurately capture the nuances and variations inherent in dialectal Arabic.

\subsection{Intent Detection by LLM Prompting}\label{intent_detection_by_prompting}
This approach leverages the capabilities of LLMs such as ChatGPT-3.5 \citep{chatgpt}, GPT-4 \citep{gpt4}, and JAIS 13B Chat \citep{jais} to classify customer intents through a strategic prompting methodology. This method provides the LLM with detailed context, its role as a classifier, and an extensive list of 24 predefined intents alongside their descriptions.
The prompt is structured as follows:
\begin{quote}
\itshape
[[ Context: You are an advanced banking chatbot designed for a Moroccan bank, equipped to assist customers with a range of inquiries and services related to banking. Your capabilities extend from handling basic account management to addressing complex service requests. Your primary objective is to accurately discern the customer's intent from their utterances, using the list of predefined intents to provide relevant assistance or guide them to the appropriate service channel.

Here is the list of all intents and their meanings:
\begin{itemize}
    \item activate\_my\_card: Initiate the use of a new banking card.
\item age\_limit: Inquire about the minimum age requirement for a service.
\item cancel\_order: Request to cancel a previously placed order.
...
\item oodoos: An intent not in the list of intents and not related to banking, like asking the distance between the Earth and the Moon.
\end{itemize}
When you receive the 5 utterances from a customer, analyze the content to determine the most applicable intents. Consider the context of banking practices in Morocco, including services and customer expectations.

Instructions:
1. Read the customer's utterances carefully.
2. Identify the most relevant intent for each utterance from the predefined list.
3. Return the detected intents in JSON format for easy parsing:
```
\{"intents": ["intent1", "intent2", "intent3", "intent4", "intent5"]\}
```

Make sure to return only one intent for each utterance. Select the intent that best matches the customer's query or service need for each of the five utterances. If an utterance does not fit any predefined intents or falls outside the banking domain, use "oodoos" for unrelated queries and "idoos" for banking-related queries not listed among the predefined intents.]]
\end{quote}
Given the complexity and the need for efficiency, multiple utterances were classified within the same prompt rather than individually, optimizing both cost and computational resources.
For JAIS 13B Chat, the approach was similar, with the adaptation being the translation of the prompt into Arabic and adjusting the formatting to suit the specifics of JAIS prompting schema.

\subsection{Intent Detection Through NMT Pipeline}\label{translation_pipeline}

One effective strategy that balances intuition and competitiveness is to employ a neural machine translation (NMT) pipeline. This approach entails translating text from a low-resource language (LRL) to a high-resource language (HRL) before executing natural language processing tasks. Research by \citep{Song:23} demonstrated that integrating a translation pipeline can yield high-quality NLP outcomes for low-resource languages. In our specific scenario, we could translate Darija queries into English and subsequently apply English-based Intent Detection models for further analysis.

\section{Experiments and Results}
In this section, we present a comprehensive analysis of our experimental studies, conducted to assess the effectiveness of the three distinct approaches previously discussed. Specifically, in subsection 5.1, we delve into the nuances of fine-tuning BERT-like models. This subsection is bifurcated into an examination of both the application of zero-shot learning techniques and the comprehensive fine-tuning process. Subsequently, subsection 5.2 is dedicated to exploring the Retrieval-based Intent Detection method, wherein we benchmark the various models listed to evaluate their performance. In subsection 5.3, we shift our focus to investigating intent detection through the lens of LLM Prompting techniques, offering a fresh perspective on this approach. The culmination of our experimental journey is presented in subsection 5.4, where we discuss the outcomes of our investigations, aiming to extract meaningful insights and implications from the gathered data. To provide a robust evaluation of model efficacy, we report performance metrics on the test set, including macro F1 scores, precision, and recall, thereby ensuring a holistic assessment of each model's capabilities.

\subsection{BERT-like models Finetuning Results} \label{finetuning_results}
\textbf{Zero-Shot Cross-Lingual Transfer Learning:}
This section explores the efficacy of zero-shot cross-lingual transfer learning on MSA and Darija, employing the XLM-Roberta model. We conduct experiments to assess the performance of this approach using the DarijaBanking English + French training dataset. Initial results revealed an F1-score of 74.06 for MSA and 47.61 for Darija (Table \ref{zeroshot}). Further experimentation, which included fine-tuning XLM-R with an enriched training dataset comprising English, French, and MSA, yielded improved F1-scores of 93.10 for MSA and 80.76 for Darija. The most comprehensive training set, encompassing English, French, Arabic, and Darija, allowed XLM-R to achieve F1-scores of 95 for MSA and 93.64 for Darija. These findings underscore the challenges faced by multilingual pre-trained models in accurately capturing the nuances of MSA and dialectical Arabic, highlighting the necessity for dedicated data annotations in these languages.

% \usepackage{graphicx}
% \usepackage{multirow}
% \usepackage{graphicx}
\begin{table*}[!htb]
\centering
\resizebox{\textwidth}{!}{%
\begin{tabular}{|l|llll|llll|ll|}
\hline
\multicolumn{1}{|c|}{\multirow{2}{*}{\textbf{\begin{tabular}[c]{@{}c@{}}Language splits of DarijaBanking\\ dataset used for training\end{tabular}}}} &
  \multicolumn{4}{c|}{\textbf{MSA split of DarijaBanking test set}} &
  \multicolumn{4}{c|}{\textbf{Darija split of DarijaBanking test set}} &
  \multicolumn{2}{l|}{\textbf{Gain in F1 Score}} \\ \cline{2-11} 
\multicolumn{1}{|c|}{} &
  \multicolumn{1}{c|}{\textbf{Accuracy}} &
  \multicolumn{1}{c|}{\textbf{Precision}} &
  \multicolumn{1}{c|}{\textbf{Recall}} &
  \multicolumn{1}{c|}{\textbf{F1\_score}} &
  \multicolumn{1}{c|}{\textbf{Accuracy}} &
  \multicolumn{1}{c|}{\textbf{Precision}} &
  \multicolumn{1}{c|}{\textbf{Recall}} &
  \multicolumn{1}{c|}{\textbf{F1\_score}} &
  \multicolumn{1}{l|}{\textbf{MSA}} &
  \textbf{Darija} \\ \hline
English, French, Arabic, Darija &
  \multicolumn{1}{l|}{95,30} &
  \multicolumn{1}{l|}{95,18} &
  \multicolumn{1}{l|}{94,94} &
  94,98 &
  \multicolumn{1}{l|}{94,20} &
  \multicolumn{1}{l|}{94,00} &
  \multicolumn{1}{l|}{93,52} &
  93,64 &
  \multicolumn{1}{l|}{+1.88} &
  +12.88 \\ \hline
English, French, Arabic &
  \multicolumn{1}{l|}{93,65} &
  \multicolumn{1}{l|}{93,29} &
  \multicolumn{1}{l|}{93,27} &
  93,10 &
  \multicolumn{1}{l|}{81,22} &
  \multicolumn{1}{l|}{83,94} &
  \multicolumn{1}{l|}{80,44} &
  80,76 &
  \multicolumn{1}{l|}{+19.04} &
  +33.15 \\ \hline
English, French &
  \multicolumn{1}{l|}{75,69} &
  \multicolumn{1}{l|}{82,88} &
  \multicolumn{1}{l|}{75,17} &
  74,06 &
  \multicolumn{1}{l|}{49,72} &
  \multicolumn{1}{l|}{67,43} &
  \multicolumn{1}{l|}{50,15} &
  47,61 &
  \multicolumn{2}{c|}{-} \\ \hline
\end{tabular}%
}
\caption{Performance of XLM-Roberta Zero-Shot Learning: Gains from Sequential Language Integration in the DarijaBanking Dataset}
\label{zeroshot}
\end{table*}

\textbf{Comprehensive Fine-tuning of Pre-Trained Transformers:}
In the preceding section, we observed that multilingual pre-trained transformers exhibit suboptimal performance on both MSA and Darija, with Darija presenting particular challenges. This section extends our evaluation to a range of Arabic pre-trained transformer models that were previously introduced in section \ref{intent_detection_by_finetuning}, including XLM-R, using the DarijaBanking dataset. We present the results in Table \ref{finetune_bench}, ranking the models by their F1-scores on the Darija test set.
Notably, Arabertv02-twitter emerges as the top-performing model, achieving impressive F1-scores of 95.55 for MSA and 97.87 for Darija. The model's efficiency is particularly noteworthy given its relatively small size, which allows for deployment on standard CPUs without compromising performance.

% \usepackage{multirow}
% \usepackage{graphicx}
\begin{table*}[!htb]
\centering
\resizebox{\textwidth}{!}{%
\begin{tabular}{|l|llll|llll|}
\hline
\multicolumn{1}{|c|}{\multirow{2}{*}{\textbf{\begin{tabular}[c]{@{}c@{}}Pre-trained Language\\ Model HuggingFace ID\end{tabular}}}} &
  \multicolumn{4}{c|}{\textbf{MSA split of DarijaBanking test set}} &
  \multicolumn{4}{c|}{\textbf{Darija split of DarijaBanking test set}} \\ \cline{2-9} 
\multicolumn{1}{|c|}{} &
  \multicolumn{1}{c|}{\textbf{Accuracy}} &
  \multicolumn{1}{c|}{\textbf{Precision}} &
  \multicolumn{1}{c|}{\textbf{Recall}} &
  \multicolumn{1}{c|}{\textbf{F1\_score}} &
  \multicolumn{1}{c|}{\textbf{Accuracy}} &
  \multicolumn{1}{c|}{\textbf{Precision}} &
  \multicolumn{1}{c|}{\textbf{Recall}} &
  \multicolumn{1}{c|}{\textbf{F1\_score}} \\ \hline
\textbf{aubmindlab/bert-base-arabertv02-twitter} &
  \multicolumn{1}{l|}{\textbf{95,86}} &
  \multicolumn{1}{l|}{\textbf{95,76}} &
  \multicolumn{1}{l|}{\textbf{95,50}} &
  \textbf{95,55} &
  \multicolumn{1}{l|}{\textbf{98,07}} &
  \multicolumn{1}{l|}{\textbf{97,98}} &
  \multicolumn{1}{l|}{\textbf{97,86}} &
  \textbf{97,87} \\ \hline
UBC-NLP/MARBERTv2 &
  \multicolumn{1}{l|}{95,30} &
  \multicolumn{1}{l|}{95,31} &
  \multicolumn{1}{l|}{95,08} &
  95,01 &
  \multicolumn{1}{l|}{97,51} &
  \multicolumn{1}{l|}{97,58} &
  \multicolumn{1}{l|}{97,24} &
  97,30 \\ \hline
UBC-NLP/ARBERT &
  \multicolumn{1}{l|}{95,03} &
  \multicolumn{1}{l|}{95,03} &
  \multicolumn{1}{l|}{94,79} &
  94,63 &
  \multicolumn{1}{l|}{97,24} &
  \multicolumn{1}{l|}{97,31} &
  \multicolumn{1}{l|}{96,98} &
  97,04 \\ \hline
UBC-NLP/MARBERT &
  \multicolumn{1}{l|}{94,75} &
  \multicolumn{1}{l|}{94,84} &
  \multicolumn{1}{l|}{94,42} &
  94,42 &
  \multicolumn{1}{l|}{97,24} &
  \multicolumn{1}{l|}{97,04} &
  \multicolumn{1}{l|}{96,97} &
  96,95 \\ \hline
qarib/bert-base-qarib &
  \multicolumn{1}{l|}{94,75} &
  \multicolumn{1}{l|}{94,84} &
  \multicolumn{1}{l|}{94,58} &
  94,56 &
  \multicolumn{1}{l|}{96,69} &
  \multicolumn{1}{l|}{96,65} &
  \multicolumn{1}{l|}{96,33} &
  96,41 \\ \hline
UBC-NLP/ARBERTv2 &
  \multicolumn{1}{l|}{\textbf{96,69}} &
  \multicolumn{1}{l|}{\textbf{96,63}} &
  \multicolumn{1}{l|}{\textbf{96,48}} &
  \textbf{96,38} &
  \multicolumn{1}{l|}{96,69} &
  \multicolumn{1}{l|}{96,55} &
  \multicolumn{1}{l|}{96,40} &
  96,41 \\ \hline
CAMeL-Lab/bert-base-arabic-camelbert-da &
  \multicolumn{1}{l|}{94,75} &
  \multicolumn{1}{l|}{94,59} &
  \multicolumn{1}{l|}{94,47} &
  94,40 &
  \multicolumn{1}{l|}{95,86} &
  \multicolumn{1}{l|}{95,62} &
  \multicolumn{1}{l|}{95,42} &
  95,44 \\ \hline
SI2M-Lab/DarijaBERT &
  \multicolumn{1}{l|}{90,61} &
  \multicolumn{1}{l|}{90,39} &
  \multicolumn{1}{l|}{90,06} &
  89,88 &
  \multicolumn{1}{l|}{95,30} &
  \multicolumn{1}{l|}{95,34} &
  \multicolumn{1}{l|}{94,92} &
  94,97 \\ \hline
CAMeL-Lab/bert-base-arabic-camelbert-mix &
  \multicolumn{1}{l|}{94,75} &
  \multicolumn{1}{l|}{94,66} &
  \multicolumn{1}{l|}{94,48} &
  94,43 &
  \multicolumn{1}{l|}{95,03} &
  \multicolumn{1}{l|}{95,02} &
  \multicolumn{1}{l|}{94,47} &
  94,64 \\ \hline
alger-ia/dziribert &
  \multicolumn{1}{l|}{93,09} &
  \multicolumn{1}{l|}{93,01} &
  \multicolumn{1}{l|}{92,51} &
  92,47 &
  \multicolumn{1}{l|}{94,48} &
  \multicolumn{1}{l|}{94,55} &
  \multicolumn{1}{l|}{93,96} &
  94,06 \\ \hline
FacebookAI/xlm-roberta-base &
  \multicolumn{1}{l|}{95,30} &
  \multicolumn{1}{l|}{95,18} &
  \multicolumn{1}{l|}{94,94} &
  94,98 &
  \multicolumn{1}{l|}{94,20} &
  \multicolumn{1}{l|}{94,00} &
  \multicolumn{1}{l|}{93,52} &
  93,64 \\ \hline
\end{tabular}%
}
\caption{Performance of various pre-trained transformers on DarijaBanking}
\label{finetune_bench}
\end{table*}

\subsection{Retrieval-based Intent Detection Results}

In this section, we conduct a comprehensive evaluation of the retrieval-based intent detection approach, leveraging the capabilities of pre-trained text embedding models as detailed in Section \ref{intent_detection_by_retrieval}, with a focus on their performance using the DarijaBanking dataset. The assessment's outcomes are systematically presented in Table \ref{retrieval_bench}, where models are ordered according to their F1 scores derived from the Darija test set.
Among the models evaluated, the "\texttt{text-embedding-3-large}" model by OpenAI stands out in the closed-source category, along with Microsoft's "\texttt{multilingual-e5-base}" in the open-source domain. The former demonstrates exceptional performance with F1 scores of 90.70 for MSA and 88.44 for Darija. The latter, while slightly trailing with F1 scores of 88.91 for MSA and 86.89 for Darija, offers significant advantages in terms of efficiency and deployability. Its smaller size enables deployment on standard CPUs without a loss in performance, and its open-source nature facilitates in-house deployment.

\begin{table*}[!htb]
\centering
\resizebox{\textwidth}{!}{%
\begin{tabular}{|l|llll|llll|}
\hline
\multicolumn{1}{|c|}{\multirow{2}{*}{\textbf{\begin{tabular}[c]{@{}c@{}}Pre-trained Language\\ Model HuggingFace ID\end{tabular}}}} &
  \multicolumn{4}{c|}{\textbf{MSA split of DarijaBanking test set}} &
  \multicolumn{4}{c|}{\textbf{Darija split of DarijaBanking test set}} \\ \cline{2-9} 
\multicolumn{1}{|c|}{} &
  \multicolumn{1}{c|}{\textbf{Accuracy}} &
  \multicolumn{1}{c|}{\textbf{Precision}} &
  \multicolumn{1}{c|}{\textbf{Recall}} &
  \multicolumn{1}{c|}{\textbf{F1\_score}} &
  \multicolumn{1}{c|}{\textbf{Accuracy}} &
  \multicolumn{1}{c|}{\textbf{Precision}} &
  \multicolumn{1}{c|}{\textbf{Recall}} &
  \multicolumn{1}{c|}{\textbf{F1\_score}} \\ \hline
\textbf{openai:   "text-embedding-3-large"} &
  \multicolumn{1}{l|}{\textbf{91,44}} &
  \multicolumn{1}{l|}{\textbf{91,39}} &
  \multicolumn{1}{l|}{\textbf{90,79}} &
  \textbf{90,70} &
  \multicolumn{1}{l|}{\textbf{89,50}} &
  \multicolumn{1}{l|}{\textbf{88,83}} &
  \multicolumn{1}{l|}{\textbf{88,62}} &
  \textbf{88,44} \\ \hline
intfloat/multilingual-e5-base &
  \multicolumn{1}{l|}{89,78} &
  \multicolumn{1}{l|}{90,50} &
  \multicolumn{1}{l|}{88,90} &
  88,91 &
  \multicolumn{1}{l|}{87,85} &
  \multicolumn{1}{l|}{87,90} &
  \multicolumn{1}{l|}{87,02} &
  86,89 \\ \hline
sentence-transformers/LaBSE &
  \multicolumn{1}{l|}{88,40} &
  \multicolumn{1}{l|}{88,52} &
  \multicolumn{1}{l|}{87,32} &
  87,44 &
  \multicolumn{1}{l|}{87,29} &
  \multicolumn{1}{l|}{86,91} &
  \multicolumn{1}{l|}{86,34} &
  86,38 \\ \hline
LASER &
  \multicolumn{1}{l|}{85,91} &
  \multicolumn{1}{l|}{86,77} &
  \multicolumn{1}{l|}{84,90} &
  84,86 &
  \multicolumn{1}{l|}{85,36} &
  \multicolumn{1}{l|}{84,78} &
  \multicolumn{1}{l|}{84,08} &
  83,85 \\ \hline
distiluse-base-multilingual-cased-v1 &
  \multicolumn{1}{l|}{85,08} &
  \multicolumn{1}{l|}{85,25} &
  \multicolumn{1}{l|}{83,88} &
  84,18 &
  \multicolumn{1}{l|}{84,25} &
  \multicolumn{1}{l|}{83,98} &
  \multicolumn{1}{l|}{82,98} &
  82,98 \\ \hline
  %sentence-transformers/\\paraphrase-multilingual-mpnet-base-v2 &
\makecell[l]{sentence-transformers/\\paraphrase-multilingual-mpnet-base-v2} &
  \multicolumn{1}{l|}{90,61} &
  \multicolumn{1}{l|}{91,04} &
  \multicolumn{1}{l|}{89,93} &
  90,14 &
  \multicolumn{1}{l|}{80,94} &
  \multicolumn{1}{l|}{80,75} &
  \multicolumn{1}{l|}{79,79} &
  79,70 \\ \hline
  %sentence-transformers/\\paraphrase-multilingual-MiniLM-L12-v2 &
  \makecell[l]{sentence-transformers/\\paraphrase-multilingual-MiniLM-L12-v2} &
  \multicolumn{1}{l|}{88,40} &
  \multicolumn{1}{l|}{88,24} &
  \multicolumn{1}{l|}{87,52} &
  87,47 &
  \multicolumn{1}{l|}{77,07} &
  \multicolumn{1}{l|}{76,59} &
  \multicolumn{1}{l|}{75,67} &
  75,62 \\ \hline
\end{tabular}%
}
\caption{Performance of various pre-trained Retrievers as Intent Detectors on DarijaBanking}
\label{retrieval_bench}
\end{table*}

\subsection{Intent Detection by LLM Prompting Results}

In this section, we explore the application of LLMs for intent detection, with a particular focus on their performance within the context of the DarijaBanking dataset.
Despite the sophisticated linguistic processing and generative capabilities of these models, their performance in the task of intent detection within the DarijaBanking dataset leaves much to be desired. ChatGPT-3.5, in particular, showcases a significant gap in effectiveness, delivering results that fall short of expectations. GPT-4, although slightly more proficient, still only achieves what can best be described as mediocre performance. This outcome is notably surprising, considering GPT-4's advanced language understanding and generation abilities, including its application to languages and dialects as complex as Moroccan Darija.
The performance of the Jais-13B model further complicates the landscape. This model demonstrates a notable difficulty in accurately aligning with the predefined set of 24 intents, often misclassifying or completely missing the correct intent. This inconsistency underscores the limitations of Jais-13B as an intent classifier despite its potential advantages in other generative applications. The evidence suggests that while Jais-13B may excel in content generation, its utility as a classifier in intent detection tasks, especially those involving the DarijaBanking dataset, is limited.
These findings indicate that the general-purpose nature of LLMs might not be ideally suited for specific classification tasks such as intent detection, particularly when dealing with languages or dialects with less online presence and resources. The study underscores the necessity for a more nuanced approach, suggesting that developing and fine-tuning smaller, domain-specific language models could offer a more effective solution, as shown in subsection \ref{finetuning_results}. Table \ref{llm_bench} reports the results obtained

% \usepackage{multirow}
% \usepackage{graphicx}
\begin{table*}[!htb]
\centering
\resizebox{\textwidth}{!}{%
\begin{tabular}{|l|llll|llll|}
\hline
\multicolumn{1}{|c|}{\multirow{2}{*}{\textbf{\begin{tabular}[c]{@{}c@{}}Pre-trained Language\\ Model HuggingFace ID\end{tabular}}}} &
  \multicolumn{4}{c|}{\textbf{MSA split of DarijaBanking test set}} &
  \multicolumn{4}{c|}{\textbf{Darija split of DarijaBanking test set}} \\ \cline{2-9} 
\multicolumn{1}{|c|}{} &
  \multicolumn{1}{c|}{\textbf{Accuracy}} &
  \multicolumn{1}{c|}{\textbf{Precision}} &
  \multicolumn{1}{c|}{\textbf{Recall}} &
  \multicolumn{1}{c|}{\textbf{F1\_score}} &
  \multicolumn{1}{c|}{\textbf{Accuracy}} &
  \multicolumn{1}{c|}{\textbf{Precision}} &
  \multicolumn{1}{c|}{\textbf{Recall}} &
  \multicolumn{1}{c|}{\textbf{F1\_score}} \\ \hline
\textbf{gpt-4} &
  \multicolumn{1}{l|}{\textbf{84,42}} &
  \multicolumn{1}{l|}{\textbf{90,2}} &
  \multicolumn{1}{l|}{\textbf{82,54}} &
  \textbf{83,21} &
  \multicolumn{1}{l|}{\textbf{76,24}} &
  \multicolumn{1}{l|}{\textbf{84,32}} &
  \multicolumn{1}{l|}{\textbf{71,87}} &
  \textbf{73,32} \\ \hline
gpt-3.5-turbo &
  \multicolumn{1}{l|}{57,29} &
  \multicolumn{1}{l|}{59,21} &
  \multicolumn{1}{l|}{56,01} &
  56,14 &
  \multicolumn{1}{l|}{52,49} &
  \multicolumn{1}{l|}{53,38} &
  \multicolumn{1}{l|}{50,06} &
  50,52 \\ \hline
\end{tabular}%
}
\caption{Performance of LLMs as Intent Detectors on DarijaBanking}
\label{llm_bench}
\end{table*}

\subsection{Intent Detection Through NMT Pipeline}

In this part, we explore the NMT-based pipeline that leverages well-resourced English models for intent classification. The first step of this pipeline involves the automatic translation of queries from target languages, such as Darija, to English using proficient models. Subsequently, the resulting English queries are then fed into an English intent classification model. 

For English Intent Classification, we fine-tuned a \texttt{Bert-base-uncased} model on the English queries extracted from our training set, employing identical hyperparameters as the other models. 

For the evaluation phase, we employed the large multilingual \texttt{hf-seamless-m4t-large} model, recently introduced by Meta \citep{seamlessm4t}, to translate the test queries from both Darija and MSA to English. Subsequently, we computed the corpus-level BLEU score between the original English queries and their translated counterparts, resulting in a score of 0.328.

\begin{table*}[!htb]
\centering
\resizebox{\textwidth}{!}{
\begin{tblr}{
  row{1} = {c},
  cell{1}{2} = {c=4}{},
  cell{1}{6} = {c=4}{},
  cell{2}{1} = {r=2}{},
  cell{2}{2} = {c},
  cell{2}{3} = {c},
  cell{2}{4} = {c},
  cell{2}{5} = {c},
  cell{2}{6} = {c},
  cell{2}{7} = {c},
  cell{2}{8} = {c},
  cell{2}{9} = {c},
  vlines,
  hline{1-2,4} = {-}{},
  hline{3} = {2-9}{},
}
                                                                & \textbf{MSA split of DarijaBanking test set} &                    &                 &                    & \textbf{Darija split of DarijaBanking test set} &                    &                 &                    \\
{\textbf{ hf-seamless-m4t-large +}\\\textbf{Bert-base-uncased}} & \textbf{Accuracy}                            & \textbf{Precision} & \textbf{Recall} & \textbf{F1\_score} & \textbf{Accuracy}                               & \textbf{Precision} & \textbf{Recall} & \textbf{F1\_score} \\
                                                                & \textbf{83,97}                               & \textbf{85,70}     & \textbf{83,71}  & \textbf{83,70}     & \textbf{89.22}                                  & \textbf{90.70}     & \textbf{89.43}  & \textbf{89.03}     
\end{tblr}}
\caption{Results of the NMT pipeline on both Darija and MSA}
\label{tab:nmt}
\end{table*}

The results presented in Table~\ref{tab:nmt} demonstrate that the NMT pipeline-based solution outperforms other models including GPT-4 for both MSA and Darija in terms of F1-score, achieving 83.70 and 89.03, respectively.

\section{Discussion}

The exploration of three different finetuning techniques for monolingual and multilingual models, in both comprehensive and zero-shot scenarios, including retrieval-based approaches and prompting with Large Language Models, for intent detection in Moroccan Darija within the banking sector, offers critical insights and implications for both future research and practical application.

First and foremost, our study demonstrates that \textbf{relying solely on cross-lingual transfer learning might not be the most effective approach for addressing the complexities of Moroccan Darija}. The initial results reveal a significative variance in F1 scores between MSA and Darija, underscoring the need for dedicated linguistic resources tailored to dialectical Arabic. This gap significantly narrows with the comprehensive integration of languages and fine-tuning, as shown by the enhanced performance of models like Arabertv02-twitter when trained on datasets that encompass Darija. \textbf{This emphasizes the value of investing in high-quality, domain-specific data labeling to improve model accuracy and efficiency}, especially in bounded domains like banking where the complexity and size of the model can be balanced with targeted, high-quality data.

Furthermore, the findings \textbf{urge caution against an overreliance on powerful LLMs for specialized classification tasks such as intent detection}. Despite their impressive generative capabilities, models like ChatGPT-3.5 and GPT-4 exhibit limitations in accurately classifying intents within the DarijaBanking dataset. This highlights the importance of incorporating specialized classifiers for the Intent Classification Module in banking chatbots, where precision in understanding customer queries is paramount. LLMs, while beneficial in generating human-like responses and enhancing the chatbot's conversational capabilities, should complement rather than replace dedicated classifiers finetuned for specific intents.

However, the study also acknowledges the \textbf{economic and logistical constraints of data labeling}, recognizing that it can be prohibitively expensive or resource-intensive for some organizations. \textbf{In such scenarios, retrieval-based approaches using pre-trained text embeddings, such as OpenAI's "text-embedding-3-large" or the "multilingual-e5-base" model, offer a viable alternative}. These models demonstrate respectable performance and efficiency, providing a practical solution for intent detection that balances cost and accuracy.

Despite these advancements, it is important to acknowledge the limitations of the study.  \textbf{The intent list used is not exhaustive}, and real-world applications will likely require further adjustments to accommodate the full spectrum of banking queries. Moreover, the current approach \textbf{does not account for the contextual and multi-intent nature of real-world conversations}, which could provide valuable signals for more accurate intent classification.

\section{Conclusion}

In conclusion, this paper introduces DarijaBanking, an innovative dataset designed to enhance intent detection within the banking sector for both Moroccan Darija and MSA. Through the adaptation and refinement of English banking datasets, DarijaBanking emerges as a vital tool for nuanced language processing, comprising over 3,600 queries across 24 intent classes. Our analysis, spanning model fine-tuning, zero-shot learning, retrieval-based techniques, and Large Language Model prompting, highlights the critical need for tailored approaches in processing dialect-specific languages and underscores the effectiveness of ArabertV0.2 when finetuned on this dataset.

The research emphasizes the importance of domain-specific classifiers and the limitations of relying solely on general-purpose Large Language Models for precise intent detection. It also presents retrieval-based approaches as practical, cost-effective alternatives for scenarios where data labeling poses significant economic and logistical challenges. These approaches provide a pragmatic balance between performance and resource allocation, facilitating the advancement of AI-driven solutions in settings that are linguistically diverse and resource-limited.

However, we recognize the limitations of our study, including the non-exhaustive nature of our intent list and the lack of consideration for the contextual and multi-intent dynamics of real-world interactions. These aspects offer avenues for future research to further refine and enhance the application of AI in understanding and servicing the diverse needs of banking customers.

By contributing to the development of resources like DarijaBanking, this paper aims to support the broader goal of making AI technologies more adaptable and effective across various linguistic contexts. In doing so, we hope to inspire continued efforts towards creating more inclusive digital banking solutions and advancing the field of NLP.

\bibliography{custom}

\begin{thebibliography}{51}
\expandafter\ifx\csname natexlab\endcsname\relax\def\natexlab#1{#1}\fi

\bibitem[{Abdelali(2021)}]{qarib}
Ahmed et~al. Abdelali. 2021.
\newblock Pre-training bert on arabic tweets: Practical considerations.
\newblock \emph{arXiv preprint arXiv:2102.10684}.

\bibitem[{Abdul-Mageed(2021)}]{arbert}
Muhammad et~al. Abdul-Mageed. 2021.
\newblock \href {https://doi.org/10.18653/v1/2021.acl-long.551} {{ARBERT} {\&} {MARBERT}: Deep bidirectional transformers for {A}rabic}.
\newblock In \emph{Proceedings of the 59th Annual Meeting of the Association for Computational Linguistics and the 11th International Joint Conference on Natural Language Processing (Volume 1: Long Papers)}, pages 7088--7105, Online. Association for Computational Linguistics.

\bibitem[{Adamopoulou(2020)}]{adamopoulou}
Eleni et~al. Adamopoulou. 2020.
\newblock An overview of chatbot technology.
\newblock In \emph{Artificial Intelligence Applications and Innovations}, pages 373--383, Cham. Springer International Publishing.

\bibitem[{Ahmed(2022)}]{Ahmed_et_al_2022}
Arfan et~al. Ahmed. 2022.
\newblock Arabic chatbot technologies: A scoping review.
\newblock \emph{Computer Methods and Programs in Biomedicine Update}, 2:100057.

\bibitem[{Algotiml(2019)}]{Algotiml_et_al_2019}
Bushra et~al. Algotiml. 2019.
\newblock Arabic tweet-act: Speech act recognition for arabic asynchronous conversations.
\newblock In \emph{Proceedings of the Fourth Arabic Natural Language Processing Workshop}, pages 183--191, Florence, Italy. Association for Computational Linguistics.

\bibitem[{Antoun(2020)}]{arabert}
Wissamet~al. Antoun. 2020.
\newblock \href {https://aclanthology.org/2020.osact-1.2} {{A}ra{BERT}: Transformer-based model for {A}rabic language understanding}.
\newblock In \emph{Proceedings of the 4th Workshop on Open-Source Arabic Corpora and Processing Tools, with a Shared Task on Offensive Language Detection}, pages 9--15, Marseille, France. European Language Resource Association.

\bibitem[{Basu(2022)}]{Basu_et_al_2022}
Samyadeep et~al. Basu. 2022.
\newblock Strategies to improve few-shot learning for intent classification and slot-filling.
\newblock In \emph{Proceedings of the Workshop on Structured and Unstructured Knowledge Integration (SUKI)}, pages 17--25. Association for Computational Linguistics.

\bibitem[{Bilah(2022)}]{Bilah_et_al_2022}
Chiva Olivia et~al. Bilah. 2022.
\newblock Intent detection on indonesian text using convolutional neural network.
\newblock In \emph{2022 IEEE International Conference on Cybernetics and Computational Intelligence (CyberneticsCom)}, pages 174--178.

\bibitem[{Boujou(2021)}]{boujou}
ElMehdi et~al. Boujou. 2021.
\newblock An open access nlp dataset for arabic dialects: Data collection, labeling, and model construction.
\newblock \emph{arXiv preprint arXiv:2102.11000}.

\bibitem[{Casanueva(2020)}]{banking77}
I{\~n}igo et~al. Casanueva. 2020.
\newblock Efficient intent detection with dual sentence encoders.
\newblock In \emph{Proceedings of the 2nd Workshop on Natural Language Processing for Conversational AI}, pages 38--45, Online. Association for Computational Linguistics.

\bibitem[{Conneau(2020)}]{conneau-etal-2020-unsupervised}
Alexis et~al. Conneau. 2020.
\newblock \href {https://doi.org/10.18653/v1/2020.acl-main.747} {Unsupervised cross-lingual representation learning at scale}.
\newblock In \emph{Proceedings of the 58th Annual Meeting of the Association for Computational Linguistics}, pages 8440--8451, Online. Association for Computational Linguistics.

\bibitem[{Darwish(2021)}]{Darwish_et_al_2021}
Kareem et~al. Darwish. 2021.
\newblock A panoramic survey of natural language processing in the arab worlds.
\newblock \emph{Commun. ACM}, 64(4):72--81.

\bibitem[{Devlin(2019)}]{devlin-etal-2019-bert}
Jacob et~al. Devlin. 2019.
\newblock \href {https://doi.org/10.18653/v1/N19-1423} {{BERT}: Pre-training of deep bidirectional transformers for language understanding}.
\newblock In \emph{Proceedings of the 2019 Conference of the North {A}merican Chapter of the Association for Computational Linguistics: Human Language Technologies, Volume 1 (Long and Short Papers)}, pages 4171--4186, Minneapolis, Minnesota. Association for Computational Linguistics.

\bibitem[{El~Mekki(2021)}]{mekki}
Abdellah et~al. El~Mekki. 2021.
\newblock Bert-based multi-task model for country and province level msa and dialectal arabic identification.
\newblock In \emph{Proceedings of the sixth Arabic natural language processing workshop}, pages 271--275.

\bibitem[{El~Mekki(2022)}]{adasl}
Abdellah et~al. El~Mekki. 2022.
\newblock Adasl: an unsupervised domain adaptation framework for arabic multi-dialectal sequence labeling.
\newblock \emph{Information Processing \& Management}, 59(4):102964.

\bibitem[{Elmadany(2018)}]{Elmadany_et_al_2018}
AbdelRahim et~al. Elmadany. 2018.
\newblock Arsas: An arabic speech-act and sentiment corpus of tweets.
\newblock \emph{OSACT}, 3:20.

\bibitem[{Essefar(2023)}]{essefar}
Kabil et~al. Essefar. 2023.
\newblock Omcd: Offensive moroccan comments dataset.
\newblock \emph{Language Resources and Evaluation}, pages 1--21.

\bibitem[{et~al.(2022)}]{e5}
Liang~Wang et~al. 2022.
\newblock \href {https://api.semanticscholar.org/CorpusID:254366618} {Text embeddings by weakly-supervised contrastive pre-training}.
\newblock \emph{ArXiv}, abs/2212.03533.

\bibitem[{et~al.(2019)}]{LASER}
Mikel~Artetxe et~al. 2019.
\newblock \href {https://doi.org/10.1162/tacl_a_00288} {Massively multilingual sentence embeddings for zero-shot cross-lingual transfer and beyond}.
\newblock \emph{Transactions of the Association for Computational Linguistics}, 7:597--610.

\bibitem[{et~al.(2023{\natexlab{a}})}]{jais}
Neha~Sengupta et~al. 2023{\natexlab{a}}.
\newblock \href {http://arxiv.org/abs/2308.16149} {Jais and jais-chat: Arabic-centric foundation and instruction-tuned open generative large language models}.

\bibitem[{et~al.(2020)}]{ai_banking}
Yingzi~Xu et~al. 2020.
\newblock \href {https://doi.org/https://doi.org/10.1016/j.ausmj.2020.03.005} {Ai customer service: Task complexity, problem-solving ability, and usage intention}.
\newblock \emph{Australasian Marketing Journal (AMJ)}, 28(4):189--199.

\bibitem[{et~al.(2023{\natexlab{b}})}]{agents}
Zhiheng~Xi et~al. 2023{\natexlab{b}}.
\newblock \href {http://arxiv.org/abs/2309.07864} {The rise and potential of large language model based agents: A survey}.

\bibitem[{Feng(2020)}]{labse}
Fangxiaoyu et~al. Feng. 2020.
\newblock \href {https://doi.org/10.48550/ARXIV.2007.01852} {Language-agnostic bert sentence embedding}.

\bibitem[{Fuad(2022)}]{Fuad_and_Al_Yahya_2022}
Ahlam et~al. Fuad. 2022.
\newblock Recent developments in arabic conversational ai: A literature review.
\newblock \emph{IEEE Access}, 10:23842--23859.

\bibitem[{Hijjawi(2013)}]{Hijjawi_et_al_2013}
Mohammad et~al. Hijjawi. 2013.
\newblock User’s utterance classification using machine learning for arabic conversational agents.
\newblock In \emph{2013 5th International Conference on Computer Science and Information Technology}, pages 223--232.

\bibitem[{Hijjawi(2014)}]{Hijjawi_et_al_2014}
Mohammad et~al. Hijjawi. 2014.
\newblock Arabchat: An arabic conversational agent.
\newblock In \emph{2014 6th International Conference on Computer Science and Information Technology (CSIT)}, pages 227--237.

\bibitem[{Inoue(2021)}]{camelbert}
Go~et~al. Inoue. 2021.
\newblock The interplay of variant, size, and task type in arabic pre-trained language models.
\newblock In \emph{Proceedings of the Sixth Arabic Natural Language Processing Workshop}, Kyiv, Ukraine (Online). Association for Computational Linguistics.

\bibitem[{Jarrar et~al.(2023)Jarrar, Birim, Khalilia, Erden, and Ghanem}]{arbanking}
Mustafa Jarrar, Ahmet Birim, Mohammed Khalilia, Mustafa Erden, and Sana Ghanem. 2023.
\newblock \href {https://doi.org/10.18653/v1/2023.arabicnlp-1.22} {{A}r{B}anking77: Intent detection neural model and a new dataset in modern and dialectical {A}rabic}.
\newblock In \emph{Proceedings of ArabicNLP 2023}, pages 276--287, Singapore (Hybrid). Association for Computational Linguistics.

\bibitem[{Joukhadar(2019)}]{Joukhadar_et_al_2019}
Alaa et~al. Joukhadar. 2019.
\newblock Arabic dialogue act recognition for textual chatbot systems.
\newblock In \emph{Proceedings of The First International Workshop on NLP Solutions for Under Resourced Languages (NSURL 2019) colocated with ICNLSP 2019-Short Papers}, pages 43--49.

\bibitem[{Lakkad(2018)}]{smart_banking_chatbot}
Brijesh Lakkad. 2018.
\newblock smart-banking-chatbot.
\newblock \url{https://github.com/Brijeshlakkad/smart-banking-chatbot}.

\bibitem[{Lewis(2020)}]{rag}
Patrick et~al. Lewis. 2020.
\newblock Retrieval-augmented generation for knowledge-intensive nlp tasks.
\newblock In \emph{Proceedings of the 34th International Conference on Neural Information Processing Systems}, NIPS'20, Red Hook, NY, USA. Curran Associates Inc.

\bibitem[{Liu(2019)}]{roberta}
Yinhan et~al. Liu. 2019.
\newblock \href {https://doi.org/10.48550/ARXIV.1907.11692} {Roberta: A robustly optimized bert pretraining approach}.

\bibitem[{Malaysha(2024)}]{arafinnlp2024}
Sanad et~al. Malaysha. 2024.
\newblock {A}ra{F}in{N}lp 2024: The first arabic financial nlp shared task.
\newblock In \emph{Proceedings of the 2nd Arabic Natural Language Processing Conference (Arabic-NLP), Part of the ACL 2024.} Association for Computational Linguistics.

\bibitem[{Mezzi(2022)}]{Mezzi_et_al_2022}
Ridha et~al. Mezzi. 2022.
\newblock Mental health intent recognition for arabic-speaking patients using the mini international neuropsychiatric interview (mini) and bert model.
\newblock \emph{Sensors}, 22(3).

\bibitem[{Nagoudi(2022)}]{turjuman}
El~Moatez Billah et~al. Nagoudi. 2022.
\newblock Turjuman: A public toolkit for neural arabic machine translation.
\newblock In \emph{Proceedings of the 5th Workshop on Open-Source Arabic Corpora and Processing Tools (OSACT5)}, Marseille, France. European Language Resource Association.

\bibitem[{Naser-Karajah(2021)}]{Naser_Karajah_et_al_2021}
Eman et~al. Naser-Karajah. 2021.
\newblock Current trends and approaches in synonyms extraction: Potential adaptation to arabic.
\newblock In \emph{Proceedings of the 2021 International Conference on Information Technology (ICIT)}, pages 428--434, Amman, Jordan. IEEE.

\bibitem[{OpenAI(2022)}]{chatgpt}
OpenAI. 2022.
\newblock \href {https://openai.com/blog/chatgpt} {Introducing chatgpt}.
\newblock Accessed: 2023-03-13.

\bibitem[{OpenAI(2023{\natexlab{a}})}]{gpt4}
OpenAI. 2023{\natexlab{a}}.
\newblock \href {https://doi.org/10.48550/ARXIV.2303.08774} {Gpt-4 technical report}.
\newblock \emph{arXiv}.

\bibitem[{OpenAI(2023{\natexlab{b}})}]{openai_embedding}
OpenAI. 2023{\natexlab{b}}.
\newblock \href {https://openai.com/blog/new-embedding-models-and-api-updates} {New embedding models and api updates}.

\bibitem[{Patel(2017)}]{banking_faq_bot}
Jay Patel. 2017.
\newblock banking-faq-bot.
\newblock \url{https://github.com/MrJay10/banking-faq-bot}.

\bibitem[{Reimers(2019)}]{sentence-bert}
Nils et~al. Reimers. 2019.
\newblock \href {https://doi.org/10.48550/ARXIV.1908.10084} {Sentence-bert: Sentence embeddings using siamese bert-networks}.

\bibitem[{Reimers(2020{\natexlab{a}})}]{distiluse}
Nils et~al. Reimers. 2020{\natexlab{a}}.
\newblock \href {https://doi.org/10.48550/ARXIV.2004.09813} {Making monolingual sentence embeddings multilingual using knowledge distillation}.

\bibitem[{Reimers(2020{\natexlab{b}})}]{distiluse_hf}
Nils et~al. Reimers. 2020{\natexlab{b}}.
\newblock \href {https://huggingface.co/sentence-transformers/distiluse-base-multilingual-cased-v1} {{sentence-transformers/distiluse-base-multilingual-cased-v1}}.

\bibitem[{Reimers(2020{\natexlab{c}})}]{minilm_hf}
Nils et~al. Reimers. 2020{\natexlab{c}}.
\newblock \href {https://huggingface.co/sentence-transformers/paraphrase-multilingual-MiniLM-L12-v2} {{sentence-transformers/paraphrase-multilingual-MiniLM-L12-v2}}.

\bibitem[{Reimers(2020{\natexlab{d}})}]{paraphrase_hf}
Nils et~al. Reimers. 2020{\natexlab{d}}.
\newblock \href {https://huggingface.co/sentence-transformers/paraphrase-multilingual-mpnet-base-v2} {{sentence-transformers/paraphrase-multilingual-mpnet-base-v2}}.

\bibitem[{{Seamless Communication}(2023)}]{seamlessm4t}
Lo\"{i}c Barrault et~al. {Seamless Communication}. 2023.
\newblock Seamlessm4t—massively multilingual \& multimodal machine translation.
\newblock \emph{ArXiv}.

\bibitem[{Shams(2019)}]{Shams_et_al_2019}
Sana et~al. Shams. 2019.
\newblock Lexical intent recognition in urdu queries using deep neural networks.
\newblock In \emph{Advances in Soft Computing}, pages 39--50, Cham. Springer International Publishing.

\bibitem[{Shams(2022)}]{Shams_and_Aslam_2022}
Sana et~al. Shams. 2022.
\newblock Improving user intent detection in urdu web queries with capsule net architectures.
\newblock \emph{Applied Sciences}, 12(22).

\bibitem[{Song(2023)}]{Song:23}
Yewei et~al. Song. 2023.
\newblock Letz translate: Low-resource machine translation for luxembourgish.
\newblock In \emph{2023 5th International Conference on Natural Language Processing (ICNLP)}, pages 165--170. IEEE.

\bibitem[{Tiedemann and Thottingal(2020)}]{opus_mt}
J{\"o}rg Tiedemann and Santhosh Thottingal. 2020.
\newblock {OPUS-MT} — {B}uilding open translation services for the {W}orld.
\newblock In \emph{Proceedings of the 22nd Annual Conferenec of the European Association for Machine Translation (EAMT)}, Lisbon, Portugal.

\bibitem[{Zhou(2022)}]{Zhou_et_al_2022}
Yunhua et~al. Zhou. 2022.
\newblock Knn-contrastive learning for out-of-domain intent classification.
\newblock In \emph{Proceedings of the 60th Annual Meeting of the Association for Computational Linguistics (Volume 1: Long Papers)}, pages 5129--5141. Association for Computational Linguistics.

\end{thebibliography}
\bibliographystyle{acl_natbib}

\end{document}